\documentclass[10pt,twocolumn,letterpaper]{article}

\usepackage{cvpr}
\usepackage{times}
\usepackage{epsfig}
\usepackage{graphicx}
\usepackage{amsmath}
\usepackage{amssymb}
\usepackage{slashbox}
\usepackage{array,multirow}
\usepackage{booktabs}
\usepackage{comment}
\usepackage{rotating}
\usepackage{verbatim}
\usepackage{stackengine}
\usepackage[utf8]{inputenc}
\usepackage{color}
\usepackage[export]{adjustbox}
\usepackage{booktabs}
\usepackage{enumitem}
\usepackage{microtype}
\usepackage{xr} 
\usepackage{tabu}

\usepackage{caption}
\usepackage[breaklinks=true,bookmarks=false]{hyperref}

\def\cvprPaperID{6102} 
\cvprfinalcopy

\usepackage{xcolor}

\begin{document}

\title{Importance Estimation for Neural Network Pruning}

\author{Pavlo Molchanov, Arun Mallya, Stephen Tyree, Iuri Frosio, Jan Kautz\\
{NVIDIA}\\
{\tt\small\{pmolchanov, amallya, styree, ifrosio, jkautz\}@nvidia.com}}

\maketitle

\begin{abstract}

Structural pruning of neural network parameters reduces computation, energy, and memory transfer costs during inference. 
We propose a novel method that estimates the contribution of a neuron (filter) to the final loss and iteratively removes those with smaller scores. 
We describe two variations of our method using the first and second-order Taylor expansions to approximate a filter's contribution.
Both methods scale consistently across any network layer without requiring per-layer sensitivity analysis and can be applied to any
kind of layer, including skip connections. 
For modern networks trained on ImageNet, we measured experimentally a high ($\textgreater 93\%$) correlation between the contribution computed by our methods and a reliable estimate of the true importance. 
Pruning with the proposed methods leads to an improvement over state-of-the-art in terms of accuracy, FLOPs, and parameter reduction.
On ResNet-101, we achieve a 40\% FLOPS reduction by removing 30\% of the parameters, with a loss of 0.02\% in the top-1 accuracy on ImageNet.
Code is available at {\small \url{https://github.com/NVlabs/Taylor_pruning}}.
\end{abstract}
\section{Introduction}
\label{sec:introduction}

Convolutional neural networks (CNNs) are widely used in today's computer vision applications. 
Scaling up the size of datasets as well as the models trained on them has been responsible for the successes of deep learning. The dramatic increase in number of layers, from 8 in AlexNet~\cite{krizhevsky2012imagenet}, to over 100 in ResNet-152~\cite{he2016identity}, has enabled deep networks to achieve better-than-human performance on the ImageNet~\cite{russakovsky2015imagenet} classification task.
Empirically, while larger networks have exhibited better performance, possibly due to the lottery ticket hypothesis~\cite{frankle2018lottery}, they have also been known to be heavily over-parameterized~\cite{zhang2016understanding}. 

\begin{figure}
    \centering
    \includegraphics[trim={0cm 0.5cm 0 0}, width=\columnwidth]{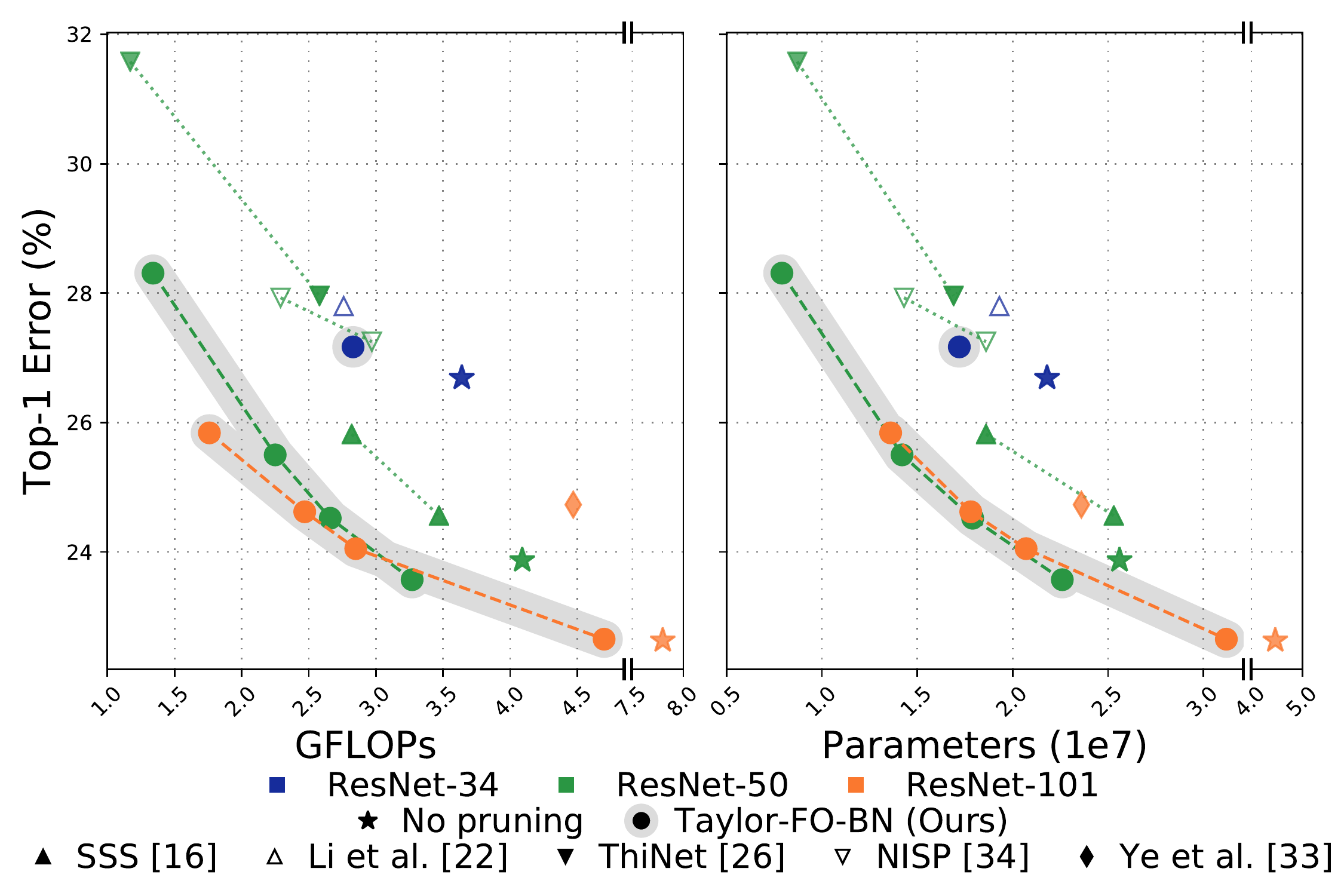}
    \vspace{0cm}
    \caption{Pruning ResNets on the ImageNet dataset. The proposed method is highlighted in gray. Bottom-left is better.}
    \label{fig:resnet_pruning_teaser}
\end{figure}

The growing size of CNNs may be incompatible with their deployment on mobile or embedded devices, with limited computational resources.
Even in the case of cloud services, prediction latency and energy consumption are important considerations.
All of these use cases will benefit greatly from the availability of more compact networks. Pruning is a common method to derive a compact network 
-- after training, some structural portion of the parameters is removed, along with its associated computations.

A variety of pruning methods have been proposed, based on greedy algorithms~\cite{luo2017thinet,yu2017nisp}, sparse regularization~\cite{lebedev2016fast,li2016pruning, ICLR2018}, and reinforcement learning~\cite{he2018adc}. Many of them rely on the belief that the magnitude of a weight and its importance are strongly correlated.
We question this belief and observe a significant gap in correlation between weight-based pruning decisions and empirically optimal one-step decisions -- a gap which our greedy criterion aims to fill.

We focus our attention on extending previously proposed methods~\cite{chauvin1989back,lecun1990optimal,ICLR2017} with a new pruning criterion and a method that iteratively removes the least important set of neurons (typically filters) from the trained model.
We define the importance as the squared change in loss induced by removing a specific filter from the network.
Since computing the exact importance is extremely expensive for large networks, we approximate it with a Taylor expansion (akin to~\cite{ICLR2017}),
resulting in a criterion computed from parameter gradients readily available during standard training. Our method is easy to implement in existing frameworks with minimal overhead.

Additional benefits of our novel criterion include: a) no hyperparameters to set, other than providing 
the desired number of neurons to prune; b) globally consistent scale of our criterion across network layers without the need for per-layer sensitivity analysis; c) a simple way of computing the criterion in parallel for all neurons without greedy layer-by-layer computation; and d) the ability to apply the method to any layer in the network, including skip connections. 
We highlight our main contributions below: 
\begin{itemize}[noitemsep,nolistsep]
    \item We propose a new method for estimating, with a little computational overhead over training, the contribution of a neuron (filter) to the final loss. 
    To do so, we use averaged gradients and weight values that are readily available during training. 
    \item We compare two variants of our method using the first and second-order Taylor expansions, respectively, against a greedy search (``oracle''), 
    and show that 
    both variants achieve state-of-the-art results, with our first-order criteria being significantly faster to compute with slightly worse accuracy.
  
    We also find that using a squared loss as a measure for contribution leads to better correlations with the oracle and better accuracy when compared to signed difference~\cite{lecun1990optimal}.
   
    Estimated Spearman correlation with the oracle on ResNets and DenseNets trained on ImageNet show significant agreement (\textgreater$93\%$), a large improvement over previous methods~\cite{huang2017data,lecun1990optimal,li2016pruning,ICLR2017,ICLR2018}, leading to improved pruning.
    \item Pruning results on a wide variety of networks trained on CIFAR-10 and ImageNet, including those with skip connections, show improvement over state-of-the-art. 
\end{itemize}

\section{Related work}
One of the ways to reduce the computational complexity of a neural network is to train a smaller model that can mimic the output of the larger model. Such an approach, termed network distillation, was proposed by Hinton~\etal~\cite{hinton2015distilling}. 

The biggest drawback of this approach is the need to define the architecture of the smaller distilled model beforehand.

Pruning -- which removes entire filters, or neurons, that make little or no contribution to the output of a trained network -- is another way to make a network smaller and faster. 
There are two forms in which structural pruning is commonly applied: a) with a predefined  per-layer pruning ratio, or b) simultaneously over all layers. The second form allows pruning to automatically find a better architecture, as demonstrated in~\cite{ICLR2019}.
An exact solution for pruning will be to minimize the $\ell_0$ norm of all neurons and remove those that are zeroed-out. However, $\ell_0$ minimization is impractical as it is non-convex, NP-hard, and requires combinatorial search. Therefore, prior work has tried to relax the optimization using Bayesian methods~\cite{louizos2017learning, neklyudov2017structured} or regularization terms. 

One of the first works that used regularization, by Hanson and Pratt~\cite{hanson1989comparing}, used weight decay along with other energy minimization functions to reduce the complexity of the neural network. At the same time, Chauvin~\cite{chauvin1989back} discovered that augmenting the loss with a positive monotonic function of the energy term 
can lead to learning a sparse solution. 

Motivated by the success of sparse coding, several methods relax $\ell_0$ minimization with $\ell_1$ or $\ell_2$ regularization, followed by soft thresholding of parameters with a predefined threshold. These methods belong to the family of Iterative Shrinkage and Thresholding Algorithms (ISTA)~\cite{daubechies2004iterative}. Han~\etal~\cite{han2015learning} applied a similar approach for removing individual weights of a neural network to obtain sparse non-regular convolutional kernels. 
Li~\etal.~\cite{li2016pruning} extended this approach 
to remove filters with small $\ell_1$ norms.

Due to the popularity of batch-normalization~\cite{ioffe2015batch} layers in recent networks~\cite{he2016identity,huang2017densely}, several approaches have been proposed for filter pruning based on batch-norm parameters~\cite{liu2017learning,ICLR2018}. These works regularize the scaling term ($\gamma$) of batch-norm layers and apply soft thresholding when value fell below a predefined threshold. Further, FLOPS-based penalties can also be included to directly reduce computational costs~\cite{gordon2018morphnet}. A more general scheme that uses an ISTA-like method on scaling factors was proposed by~\cite{huang2017data} and can be applied to any layer.

All of the above methods explicitly rely on the belief that the magnitude of the weight or neuron is strongly correlated with its importance. This belief was investigated as early as 1988 by Mozer~\cite{mozer1989skeletonization} who proposed adding a gating function after each layer to be pruned. With gate values initialized to $1$, the expectation of the negative gradient is used as an approximation for importance. Mozer noted that weights magnitude merely reflect the statistics of importance.
LeCun~\etal~\cite{lecun1990optimal} also questioned whether magnitude is a reasonable measure of neuron importance. The authors suggested using a product of the Hessian's diagonal and the squared weight as a measure of individual parameter importance, 
and demonstrated improvement over magnitude-only pruning. 

This approach assumes that after convergence, the Hessian is a positive definite matrix, meaning that removing any neuron will only increase the loss. However, due to stochasticity in training with minibatches under a limited observation set and in the presence of saddle points, there do exist neurons whose removal will decrease the loss.

Our method does not assume that the contribution of all neurons is strictly positive. Therefore, we approximate the squared difference of the loss when a neuron is removed and can do so with a first-order or second-order approximation, if the Hessian is available.

A few works have estimated neuron importance empirically.
Luo~\etal~\cite{luo2017thinet} propose to use a greedy per-layer procedure to find the subset of neurons that minimize a reconstruction loss, at a significant computational cost. Yu~\etal~\cite{yu2017nisp} estimate the importance of input features to a linear classifier and propagate their importance assuming Lipschitz continuity, requiring additional computational costs and non-trivial implementation of the feature score computation. Our proposed method is able to outperform these methods while requiring little additional computation and engineering.

Pruning methods such as~\cite{he2018adc,he2017channel,li2016pruning,luo2017thinet,yu2017nisp} require sensitivity analysis in order to estimate the pruning ratio that should be applied to particular layers. 
Molchanov~\etal~\cite{ICLR2017} assumed all layers have the same importance in feed-forward networks and proposed a normalization heuristic for global scaling. However, this method fails in networks with skip connections. Further, it computes the criterion using network activations, which increases memory requirements.
Conversely, pruning methods operating on batch-normalization \cite{gordon2018morphnet,huang2017data,liu2017learning,ICLR2018} do not require sensitivity analysis and can be applied globally. Our criterion has globally-comparable scaling by design and does not require sensitivity analysis. It can be efficiently applied to any layer in the network, including skip connections, and not only to batch-norm layers.

A few prior works have utilized pruning as a network training regularizer. Han~\etal~\cite{han2016dsd} re-initialize weights after pruning and finetune them to achieve even better accuracy than the initial model. He~\etal.~\cite{he2018progressive} extend this idea by training filters even after they were zeroed-out. While our work focuses only on removing filters from networks, it might be possible to extend it as a regularizer.

\section{Method}
\label{sec:method}

Given neural network parameters $\textbf{W}=\{w_0, w_1, ..., w_M\}$ and a dataset $\mathcal{D}=\{(x_0,y_0), (x_1,y_1), ..., (x_K,y_K)\}$ composed of input ($x_i$) and output ($y_i$) pairs, the task of training is to minimize error $E$ by solving:
\begin{equation}
\min_\textbf{W} {E}(\mathcal{D},\textbf{W}) =\min_\textbf{W}E(y|x,\textbf{W}).
\label{eq:main_optimizaiton}
\end{equation}

In the case of pruning we can include a sparsification term in the cost function to minimize the size of the model:
\begin{equation}
\min_\textbf{W} {E}(\mathcal{D},\textbf{W}) + \lambda ||\textbf{W}||_0,
\end{equation}
where $\lambda$ is a scaling coefficient and $||\cdot||_0$ is the $\ell_0$ norm which represents the number of non-zero elements.
Unfortunately there is no efficient way to minimize the $\ell_0$ norm as it is non-convex, NP-hard, and requires combinatorial search.

An alternative approach starts with the full set of parameters $\mathbf{W}$ upon convergence of the original optimization (\ref{eq:main_optimizaiton}) and gradually reduces this set by a few parameters at a time.
In this incremental setting, the decision of which parameters to remove can be made by considering the importance of each parameter individually, assuming independence of parameters. We refer to this simplified approximation to full combinatorial search as \textit{greedy first-order search}.

The importance of a parameter can be quantified by 
the error induced by removing it. Under an \textit{i.i.d.} assumption, this induced error can be measured as a squared difference of prediction errors with and without the parameter ($w_m$):
\begin{equation}
    \mathcal{I}_m = \bigg({E}(\mathcal{D},\textbf{W}) - {E}(\mathcal{D},\textbf{W}|w_m = 0)\bigg)^2.
    \label{eq:importance}
\end{equation}
Computing $\mathcal{I}_m$ for each parameter, as in (\ref{eq:importance}), is computationally expensive since it requires evaluating $M$ versions of the network, one for each removed parameter. 

We can avoid evaluating $M$ different networks by approximating $\mathcal{I}_m$ 
in the vicinity of $\textbf{W}$ by its second-order Taylor expansion:

\begin{equation}
    \mathcal{I}^{(2)}_m(\textbf{W}) = \bigg(g_m w_m  -  \frac{1}{2} w_m \textbf{H}_m \textbf{W}\bigg)^2,
\label{eq:taylor2}
\end{equation}
where $g_m = \frac{\partial {E}}{\partial w_m}$ are elements of the gradient $\textbf{g}$, $H_{i,j}=\frac{\partial^2 {E}}{\partial w_i \partial w_j}$ are elements of the Hessian $\textbf{H}$, and $\textbf{H}_m$ is its \textit{m}-th row.
An even more compact approximation is computed using the first-order expansion, which simplifies to:

\begin{equation}
    \mathcal{I}^{(1)}_m(\textbf{W}) =  \bigg( g_m w_m\bigg)^2.
\label{eq:taylor1}
\end{equation}
The importance in Eq. (\ref{eq:taylor1}) is easily computed since the gradient $\textbf{g}$ is already available from backpropagation. For the rest of this section we will primarily use the first-order approximation, however most statements also hold for the second-order approximation.
Future reference we denote the set of first-order importance approximations:
\begin{equation}
\textbf{I}^{(1)}(\textbf{W})~=~\{ \mathcal{I}^{(1)}_1(\textbf{W}),  \mathcal{I}^{(1)}_2(\textbf{W}), ..., \mathcal{I}^{(1)}_M(\textbf{W})\}.
\end{equation}

To approximate the joint importance of a structural set of parameters $\textbf{W}_\mathcal{S}$, e.g. a convolutional filter, we have two alternatives.
We can define it as \textit{a group contribution}:
\begin{equation}
\mathcal{I}^{(1)}_\mathcal{S}(\textbf{W}) \triangleq \left(\sum_{s\in S}g_{s}w_{s}\right)^2,
\label{eq:group}
\end{equation}
or, alternatively, \textit{sum the importance of the individual parameters} in the set,
\begin{equation}
\widehat{\mathcal{I}}^{(1)}_\mathcal{S}(\textbf{W}) \triangleq \sum_{s\in \mathcal{S}}\mathcal{I}^{(1)}_s(\textbf{W}) = \sum_{s\in \mathcal{S}}(g_s w_s)^2.
\label{eq:individ}
\end{equation}

For insight into these two options, and to simplify calculations, we add ``gates'' to the network, $\textbf{z} = \textbf{1}^M$, with weights equal to $1$ and dimensionality equal to the number of neurons (feature maps) $M$. Gating layers make importance score computation easier, as they: a) are not involved in optimization; b) have a constant value, therefore allowing $\textbf{W}$ to be omitted from Eq.~(\ref{eq:taylor2}-\ref{eq:individ}); and c) implicitly combine the contributions of filter weights and bias.

If a gate $z_m$ follows a neuron parameterized by weights $\textbf{W}_{s\in \mathcal{S}_m}$, then the importance approximation $\mathcal{I}^{(1)}_m$ is:
\begin{equation}
\mathcal{I}^{(1)}_m(\textbf{z}) = \bigg(\frac{\partial {E}}{\partial \textbf{z}_m}\bigg)^2 = \bigg(\sum_{s\in \mathcal{S}_m}g_{s}w_{s}\bigg)^2 = \mathcal{I}_{\mathcal{S}_m}^{(1)}(\textbf{W}),
\label{eq:gate}
\end{equation}
where $\mathcal{S}$ represents the inner dimensions needed to compute the output of the previous layer, \eg input dimension for a linear layer, or spatial and input dimensions for a convolutional layer.  We see that gate importance is equivalent to \textit{group contribution} on the parameters of the preceding layer.

Through some manipulation, we can make a connection to information theory from our proposed method.
Let's denote $\textbf{h}_m = \frac{\partial {E}}{\partial \textbf{z}_m}=\textbf{g}^T_{s\in \mathcal{S}_m}\textbf{W}_{s\in \mathcal{S}_m}$ and observe (under the assumption that, at convergence, $\mathop{\mathbb{E}}( \textbf{h}_m)^2 = 0$):
\begin{align}
    &\text{Var}(\textbf{h}_m) = \mathop{\mathbb{E}}( \textbf{h}_m^2 ) - \mathop{\mathbb{E}}( \textbf{h}_m)^2 = \textbf{I}^{(1)}(\textbf{z}),
\end{align}
where the variance is computed across observations.

If the error function $E(\cdot)$ is chosen to be the log-likelihood function, then assuming the gradient is estimated as $\textbf{h}_x=\frac{\partial \ln p(x;\textbf{z})}{\partial \textbf{z}}$, borrowing from concepts in information theory~\cite{cover2012elements}, we obtain
\begin{align}
    \mbox{Var}_x(\textbf{h}) = \mathop{\mathbb{E}}_x\big\{ \textbf{h}_x\textbf{h}_x^T\big\}=
    J(\textbf{h}),
\end{align}
where $J$ is the expected Fisher information matrix. We conclude that the variance of the gradient is the expectation of \textit{the outer product of gradients} and is equal to the \textit{expected Fisher information} matrix. Therefore, the proposed metric, $\textbf{I}^{(1)}$, can be interpreted as the variance estimate and as the diagonal of the Fisher information matrix. Similar conclusion was drawn in~\cite{theis2018faster}.

\subsection{Pruning algorithm}
Our pruning method takes a \textit{trained network} as input and prunes it during an iterative fine-tuning process with a small learning rate.
During each epoch, the following steps are repeated:
\begin{enumerate}[noitemsep,nolistsep]
    \item For each minibatch, we compute parameter gradients and update network weights by gradient descent. We also compute the importance of each neuron (or filter) using the gradient averaged over the minibatch, as described in (\ref{eq:group}) or (\ref{eq:individ}). (Or, the second-order importance estimate may be computed if the Hessian is available.)
    \item After a predefined number of minibatches, we average the importance score of each neuron (or filter) over the of minibatches, and remove the $N$ neurons with the smallest importance scores.
\end{enumerate}
Fine-tuning and pruning continue until the target number of neurons is pruned, or the maximum tolerable loss can no longer be achieved.

\subsection{Implementation details}
\noindent{\bf Hessian computation.}
Computing the full Hessian in Eq.~(\ref{eq:taylor2}) is computationally demanding, thus we use a diagonal approximation. During experiments with ImageNet we cannot compute the Hessian because of memory constraints.

\noindent{\bf Importance score accumulation.}
During training or fine-tuning with minibatches, observed gradients are combined to compute a single importance score $\hat{\textbf{I}} = {\mathbb{E}}\big\langle \textbf{I} \big\rangle$.

\noindent{\bf Importance score aggregation.}
In this work, we compute the importance of structured parameters as a sum of individual contributions defined in Eq. (\ref{eq:individ}), unless gates are used automatically compute the group contribution on the parameters from the preceding layer. Second-order methods are always computed on gates. We observed that the ``group contribution'' criterion in Eq.~(\ref{eq:group}) exhibits very low correlation with the ``true'' importance (\ref{eq:importance}) if the parameter set $\mathcal{S}$ is too large, due to expectation of gradients tending to zero at convergence.

\noindent{\bf Gate placement.} Unless otherwise stated, gates are placed immediately after a batch normalization layer to capture contributions from scaling and shifting parameters simultaneously. The first-order criterion computed for a feature map $m$ at the gate can be shown to be $\textbf{I}^{(1)}_m(\gamma_m, \beta_m)=(\gamma_m\frac{\partial {E}}{\partial \gamma_m} + \beta_m\frac{\partial {E}}{\partial \beta_m})^2$ with $\gamma$ and $\beta$ being the scale and shift parameters of the batch normalization.

\noindent{\bf Averaging importance scores over pruning iterations.} We average importance scores between pruning iterations using an exponential moving average filter (momentum) with coefficient $0.9$. 

\noindent{\bf Pruning strategy.} We found that the method performs better when we define the number of neurons to be removed, prune them in batches and fine-tune the network after that. An alternative approach is to continuously prune as long as the training or validation loss is below the threshold. The latter approach leads the optimization into local minima and final results are slightly worse. 

\noindent{\bf Number of minibatches} between pruning iterations needs be sufficient to capture statistics of the overall data. We use $10$ minibatches and a small batch size for CIFAR datasets, but a larger ($256$) batch size and $30$ minibatches for ImageNet pruning, as noted with each experiment.

\noindent{\bf Number of neurons} pruned per iteration needs to be chosen based on how correlated the neurons are to each other. We observed that a filter's contribution changes during pruning and we usually prune around $2\%$ of initial filters per iteration.  
\section{Experiments}
\label{sec:experiments}

We evaluate our method on a variety of neural network architectures on the CIFAR-10~\cite{krizhevsky2009cifar} and ImageNet~\cite{russakovsky2015imagenet} datasets. We also experiment with variations of our method to understand the best variant. Whenever we refer to \textit{Weight}, \textit{Weight magnitude} or \textit{BN scale}, we use $\ell_2$ norm.

\subsection{Results on CIFAR-10}
With the CIFAR-10 dataset, we evaluate ``\textit{oracle}'' methods and second-order methods by pruning smaller networks, including LeNet3 and variants of ResNets~\cite{he2016deep} and pre-activation ResNets~\cite{he2016identity}.

\subsubsection{LeNet3}
We start with a simple network, LeNet3, trained on the CIFAR-10 dataset to achieve $73\%$ test accuracy. The architecture of LeNet consists of $2$ convolutional and $3$ linear layers arranged in a 
\texttt{C-R-P-C-R-P-L-R-L-R-L} (C: Conv, R: ReLU, P: Pooling, L: Linear)
order with $16$, $32$, $120$, $84$, and $10$ neurons respectively. We prune the first $2$ convolutional and first $2$ linear layers without changing the output linear layer or finetuning after pruning. 

\begin{figure}[h]
\centering
\includegraphics[width=\columnwidth,trim={0 0 0 1.5cm},clip]{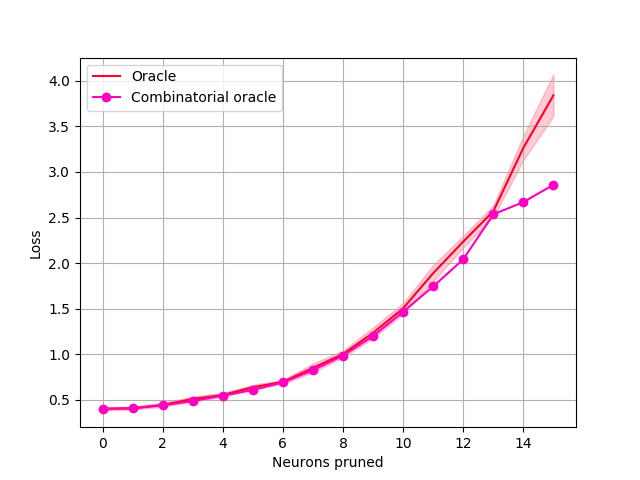}
\caption{Pruning the first layer of LeNet3 on CIFAR-10 with Combinatorial oracle and Greedy oracle. Networks remain fixed and are not fine-tuned. Results for Greedy oracle are averaged over $30$ seeds with mean and standard deviation shown. Best observed results for Combinatorial oracle for every seed are averaged.}
\label{fig:lenet_pruning_first_layer_only}
\end{figure}

\noindent{\bf Single layer pruning.}
In this setup, we only prune the first convolutional layer. This setting allows us to use the \emph{Combinatorial oracle}, the true $\ell_{0}$ minimizer: we compute the loss for all possible combinations of  $k$ neurons that can be pruned and pick the best one.
Note that this requires an exponential number of feedforward passes to evaluate -- ${n \choose k}$ per $k$, where $n$ is the number of filters and $k$ is number of filters to prune, and so is not practical for multiple layers or larger networks.
We compare against a greedy search approximation, the \emph{Greedy oracle}, that exhaustively finds the single best neuron to remove at each pruning step, repeated $k$ times.
Results shown in Fig.~\ref{fig:lenet_pruning_first_layer_only} show the loss vs. the number of neurons pruned.
We observe that the \emph{Combinatorial oracle} is not significantly better than the \emph{Greedy oracle} when pruning a small number of neurons.
Considering that the former has exponential computational complexity, in subsequent experiments we use the \emph{Greedy oracle} (referred to simply as \emph{Oracle}) as a representation of the best possible outcome.

\smallskip
\noindent{\bf All layers pruning.}
\begin{figure}[b]
\centering
\includegraphics[width=\columnwidth,trim={0 0 0 1.5cm},clip]{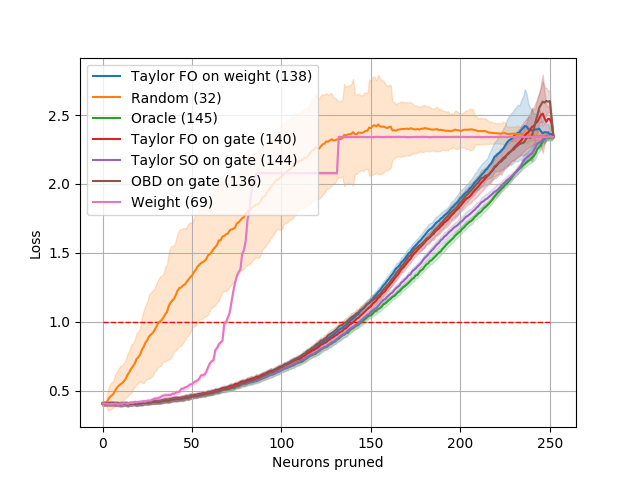}
\caption{Pruning LeNet3 on CIFAR-10 with various criteria. Network remains fixed and is not fine-tuned. Results are averaged over $50$ seeds with mean and standard deviation. The number of pruned neurons when the loss reaches $1.0$ is shown in parentheses.}
\label{fig:lenet_pruning_all}
\end{figure}
Fig.~\ref{fig:lenet_pruning_all} shows pruning results when all layers are pruned using various criteria. We refer to our methods based on the Taylor expansion as \emph{Taylor FO}/\emph{Taylor SO}, indicating the order of the approximation used, first- and second-order, respectively. 
We consider both a direct application to convolutional filter weights (``on weight'') and the use of gates following each convolutional layer (``on gate'').]
We treat linear layers as $1\times 1$ convolutions. 
In all cases, pruning removes the entire filter and its corresponding bias.
At each pruning iteration, we remove the neuron with the least importance as measured by the criterion used, and measure the loss on the training set.

Results in Fig.~\ref{fig:lenet_pruning_all} show that \textit{Oracle} pruning performs best, followed closely by the second- and first-order Taylor expansion criteria, respectively. Both first and second-order Taylor methods prune nearly the same number of neurons as the \textit{Oracle} before exceeding the loss threshold. Weight-based pruning, which removes neurons with the least $\ell_2$ norm, performs as poorly as randomly removing neurons. OBD~\cite{lecun1990optimal} performs similarly to the \textit{Oracle} and Taylor methods.

The experiments on LeNet confirm the following: (1) The greedy \emph{oracle} closely follows the pruning performance of the \emph{Combinatorial oracle} for small changes to the network, while being exponentially faster to compute. (2) Our first-order method (Taylor FO) is comparable to the second-order method (Taylor SO) in this setting.
\subsubsection{ResNet-18 }

Now we compare pruning criteria on the more complex architecture ResNet-18, 
from the pre-activation family~\cite{he2016identity}.
Each residual block has an architecture of \texttt{BN1-ReLU-conv1-BN2-ReLU-conv2}, together with a skip connection from the input to the output, repeated for a total of $8$ blocks.
Trained on CIFAR-10, ResNet-18 achieves a test accuracy of $94.79\%$. 
For pruning, we consider entire feature maps in the \texttt{conv} layers as they command the largest share of computational resources. 

In these experiments, we examine the following ways of estimating our criterion: (1) Applying it directly on convolutional filter weights, (2) Using gates placed before \texttt{BN2} and after \texttt{conv2}, and (3) Using gates placed after \texttt{BN2} and after \texttt{conv2}.
We remove $100$ neurons every $20$ minibatches, and report final results averaged over $10$ seeds.
We also compare using gradients averaged over a mini-batch and gradients obtained per data sample, the latter denoted by ``full grad'', or ``FG''. We should note that using the full gradient changes the gate formulation from computing the group contribution (Eq.~\ref{eq:group}) to the sum of individual contributions (Eq.~\ref{eq:individ}).

Table~\ref{tab:resnet_cor} presents the Spearman correlation between various pruning criteria and the greedy oracle. 
Results in the \emph{Residual block} column are averaged over all $8$ blocks.
The \emph{All layers} column includes additional layers: the first convolutional layer (not part of residual blocks), all convolutions in residual blocks, and all strided convolutions. 
We observe that placing the gate before \texttt{BN2} significantly reduces correlation -- correlation for \texttt{conv1} drops from $0.95$ to $0.28$ for Taylor FO, suggesting that the subsequent batch-normalization layer significantly affects criteria computed from the gate. We observe that the effect is less significant when the full gradient is used, however it shows smaller correlation overall with the oracle.
OBD has lower correlation than our Taylor based methods. 
The highest correlation is observed for Taylor SO, with Taylor FO following right after. As placing gates after \texttt{BN2} dramatically improves the results, this indicates that the batch-normalization layers play a key role in determining the contribution of the corresponding filter.

\begin{table}[t!]
\centering
\resizebox{\columnwidth}{!} 
{
\begin{tabular}{lccc}

\toprule
\multirow{2}{*}{Method} & \multicolumn{2}{c}{Residual block} & \multirow{2}{*}{All layers} \\
 & \texttt{conv1} & \texttt{conv2}  \\ 
\midrule
Taylor FO on conv weight & $0.726$ & $0.766$ & $0.892$ \\
Weight magnitude & $0.660$ & $0.677$ & $0.861$ \\
\midrule
\multicolumn{4}{c}{Gate after \texttt{BN2}} \\
Taylor FO & $0.955$ & $0.811$ & $0.924$ \\
Taylor SO & $\mathbf{0.968}$ & $\mathbf{0.849}$ & $\mathbf{0.957}$ \\
OBD & $0.855$ & $0.707$ & $0.901$ \\
Taylor FO - FG & $0.775$ & $0.746$ & $0.924$ \\
\midrule
\multicolumn{4}{c}{Gate before \texttt{BN2}} \\
Taylor FO  & \underline{$0.275$} & $0.811$ & \underline{$0.295$} \\
Taylor SO & \underline{$0.376$} & $0.849$ & \underline{$0.286$} \\
OBD &  \underline{$0.350$} & $0.707$ & \underline{$0.299$} \\
Taylor FO - FG & $0.642$ & $0.746$ & $0.900$ \\

\bottomrule
\end{tabular}
}
\caption{Spearman correlation of different criteria with the Oracle on CIFAR-10 with ResNet-18. (FG denotes full gradient, as described in the text).}
\label{tab:resnet_cor}
\end{table}

\begin{figure}
\centering
\includegraphics[width=\columnwidth]{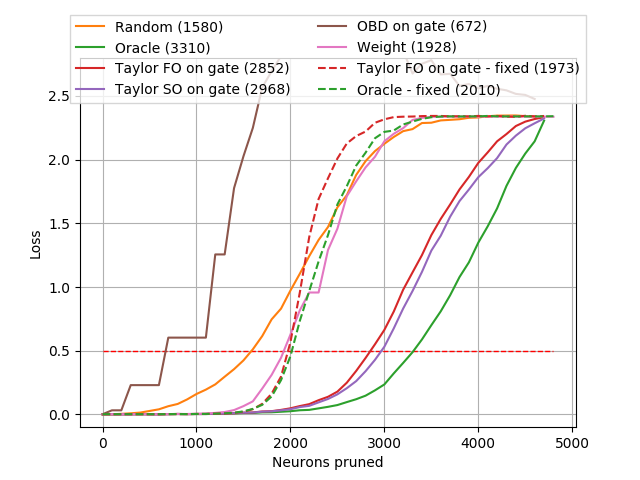}
\caption{Pruning ResNet-18 trained on CIFAR-10 without fine-tuning. The number of neurons pruned when the loss reaches $0.5$ is shown in parentheses.}
\label{fig:resnet18}
\end{figure}

Results of pruning ResNet-18 without fine-tuning are shown in Fig.~\ref{fig:resnet18}. We observe that the oracle achieves the best accuracy for a given number of pruned neurons. 
All methods, except ``-fixed'' and Random, recompute the criteria after each iterative step and can adjust to the pruned network. Oracle-fixed and Taylor FO-fixed are computed across the same number of batches as non-fixed criteria.
We notice that fixed criteria clearly perform significantly worse than oracle, emphasizing importance of reestimating the criteria after each pruning iteration, allowing the values to adjust to changes in the network architecture.

An interesting observation is that the OBD method performs poorly in spite of having a good correlation with the oracle in Table~\ref{tab:resnet_cor}. 
The reason for this discrepancy is that when we evaluate correlation with the oracle, we square estimates of OBD to make them comparable to the way the oracle was estimated. However, during pruning, we use signed values of OBD, as was prescribed in~\cite{lecun1990optimal}.
As mentioned earlier, for deep networks, the diagonal of the Hessian is not positive for all elements and removing those with negative impact results in increased network instability. Therefore, without fine-tuning, OBD is not well suited for pruning. Another important observation is that if the Hessian is available, using the Taylor SO expansion can get both better pruning and correlation.
Surprisingly, we observe no improvement in using the full gradient, probably because of the switch in contributions from group to individual.

At this stage, after experiments with the small LeNet3 network the larger ResNet-18 on the CIFAR-10 dataset, we make the following observations: 
(1) Our proposed criteria based on the Taylor expansion of the pruning loss have a very high correlation with the neuron ranking produced by the oracle. 
(2) The first- and second-order Taylor criteria are comparable. As the Taylor FO can be computed much faster with a lower memory footprint, further experiments with larger networks on ImageNet are performed using this criterion only.
\subsection{Results on ImageNet}
Here, we apply our method on the challenging task of pruning networks trained on ImageNet~\cite{russakovsky2015imagenet}, specifically the ILSVRC2012 version. 
For all experiments in this section, we use PyTorch~\cite{pytorch} and default pretrained models as a starting point for network pruning. We use standard preprocessing and augmentation: re-sizing images to have a smallest dimension of $256$, randomly cropping a $224\times 224$ patch, randomly applying horizontal flips, and normalizing images by subtracting a per-dataset mean and dividing by a per-dataset standard deviation. During testing, we use the central crop of size $224\times 224$.

\subsubsection{Neuron importance correlation study}

We compare against pruning methods that use various heuristics, such as \emph{weight magnitude}~\cite{lebedev2016fast,li2016pruning}, magnitude of the batch-norm scale, \emph{BN scale}~\cite{gordon2018morphnet, liu2017learning, ICLR2018}, and \emph{output}-based heuristics (Taylor expansion applied to layer outputs)~\cite{ICLR2017}.

We estimate the correlation between the ``real importance'' of a filter and these criteria.
Estimating real importance, or the change in loss value upon removing a neuron, requires running inference multiple times while setting each individual filter to $0$ in turn.
(Note that the oracle ranks neurons based on this value).
For ResNet-101, we pruned filters in the first $2$ convolutional layers
of every residual block.
Separately, we add gates to skip connections at the input and output of each block. 
For the VGG11-BN architecture, we replace drop-out layers with batch-norms ($0.5$ scale and $0$ shift) and fine-tune for $12$ epochs until test accuracy reaches $70.8\%$ to be comparable with~\cite{ICLR2019}.
For DenseNet201, we considered features after the batch-norm layer that follows the first $1\times 1$ convolution in every dense layer.

\begin{table}
\centering
\resizebox{\linewidth}{!} 
{
\begin{tabular}{lccccccc}
\toprule
\multirow{2}{*}{Method} & Ours &\multicolumn{3}{c}{Averaged per layer} &  \multicolumn{3}{c}{All layers} \\ 
& \small{Taylor FO}&\small{Pearson} &  \small{Spearman} &  \small{Kendall} &  \small{Pearson} &  \small{Spearman} &  \small{Kendall} \\
\midrule
\multicolumn{8}{c}{\textbf{ResNet-101}}  \\

\midrule
Gate after BN & \checkmark & $\mathbf{0.877}$ & $\mathbf{0.870}$ & $\mathbf{0.710}$ & $\mathbf{0.925}$ & $\mathbf{0.965}$ & $\mathbf{0.843}$ \\
Gate after BN - FG & \checkmark & $0.772$ & $0.817$& $0.644$ & $0.778$ & $0.944$ & $0.803$ \\
Conv weight & \checkmark & $0.719$     &    $0.740$     &    $0.570$     &    $0.780$    &     $0.874$    &     $0.698$ \\
BN scale & \checkmark &$0.703$ & $0.664$& $0.501$ & $0.792$ & $0.866$ & $0.681$ \\
BN scale & & $0.371$ & $0.405$& $0.296$ & $0.632$ & $0.807$ & $0.621$ \\
Weight magnitude& &$0.566$ & $0.651$& $0.493$ & $0.376$ & $0.587$ & $0.432$ \\
Taylor-output~\cite{ICLR2017}& & $0.520$ & $0.586$& $0.429$ & $0.381$ & $0.287$ & $0.198$ \\\hline
\multicolumn{8}{c}{Including skip connections}  \\\hline
Gate after BN & \checkmark &$0.874$     &  $0.867$     & $0.707$     & $0.806$    & $0.946$    & $0.809$ \\
Gate after BN - FG & \checkmark & $0.768$     & $0.814$     &$0.640$     &$0.725$    &$0.873$    &$0.727$ \\
\midrule
\multicolumn{8}{c}{\textbf{VGG11-BN}}  \\
\midrule
Gate after BN & \checkmark & $\mathbf{0.964}$ & $\mathbf{0.974}$ & $\mathbf{0.894}$ & $\mathbf{0.798}$ & $\mathbf{0.999}$ & $\mathbf{0.972}$ \\
Conv/Linear weight & \checkmark & $0.659$ & $0.627$& $0.507$ & $0.843$ & $0.983$ & $0.893$ \\
Gate after BN - FG& \checkmark & $0.825$ & $0.800$& $0.666$ & $0.812$ & $0.982$ & $0.887$ \\
BN scale & \checkmark &$0.751$ & $0.718$& $0.586$ & $0.634$ & $0.968$ & $0.846$ \\
BN scale && $0.474$ & $0.438$& $0.351$ & $0.031$ & $0.257$ & $0.213$ \\
Weight magnitude && $0.604$ & $0.603$& $0.474$ & $0.537$ & $0.812$ & $0.654$ \\
Taylor-output~\cite{ICLR2017} && $0.590$ & $0.581$& $0.468$ & $0.534$ & $0.968$ & $0.876$ \\
\midrule
\multicolumn{8}{c}{\textbf{DenseNet-201}}  \\
\midrule
Gate after BN & \checkmark & $\mathbf{0.825}$ & $\mathbf{0.849}$ & $\mathbf{0.740}$ & $\mathbf{0.967}$ & $\mathbf{0.932}$ & $\mathbf{0.811}$ \\
Gate after BN - FG & \checkmark & $0.891$ & $0.898$& $0.742$ & $0.764$ & $0.944$ & $0.798$ \\
Conv weight & \checkmark & $0.825$ & $0.836$& $0.659$ & $0.701$ & $0.817$ & $0.627$ \\
BN scale & \checkmark & $0.645$ & $0.645$& $0.479$ & $0.677$ & $0.434$ & $0.307$ \\
BN scale && $0.472$ & $0.471$& $0.342$ & $0.506$ & $0.597$ & $0.436$ \\
Weight magnitude && $0.725$ & $0.737$& $0.558$ & $0.300$ & $0.300$ & $0.208$ \\
Taylor-output~\cite{ICLR2017} && $0.673$ & $0.699$& $0.530$ & $0.455$ & $0.472$ & $0.333$ \\

\bottomrule
\end{tabular}
}
\caption{Correlation study of different criteria and oracle on the ImageNet dataset. Spearman and Kendall measure rank correlations. BN stands for batch-normalization, FG for full gradient.}
\label{tab:imagemet_correlation}
\end{table}

The statistical correlation between heuristics and measured importance are summarized in Table \ref{tab:imagemet_correlation}.
Correlations were measured on a subset of ImageNet consisting of a few thousand images.
We evaluated various implementations of our method, but always use the first-order Taylor expansion, denoted Taylor FO.
As previously discussed, the most promising variation uses a gate after each batch-norm layer.
The \textit{All layers} correlation columns show how well the criteria scale across layers. Our method exhibits \textgreater$93\%$ Spearman correlation for all three networks. \textit{Weight magnitude} and \textit{BN scale} have quite low correlation, suggesting that magnitude is not a good representation of importance.
\emph{Output}-based expansion proposed in \cite{ICLR2017} has high correlation on the VGG11-BN network but fails on ResNet and DenseNet architectures. Surprisingly, we observe \textgreater$92\%$ Pearson correlation for ResNet and DenseNet, showing we can almost exactly predict the change in loss for every neuron.

We are also able to study the effect of skip connections by adding a gate after the output of each residual block.  We add skip connections to the full set of filters and evaluate their correlation, denoted ``Including skip connections" in Table~\ref{tab:imagemet_correlation}. We observe high correlation of the criterion with skip connections as well.
Given this result, we adopt this methodology for pruning ResNets and remove channels from skip connections and bottleneck layers simultaneously. 
We refer to this variant of our method as \emph{Taylor-FO-BN}.

\subsubsection{Pruning and fine-tuning}
\begin{figure*}[t]
\centering
\includegraphics[width=0.97\textwidth]{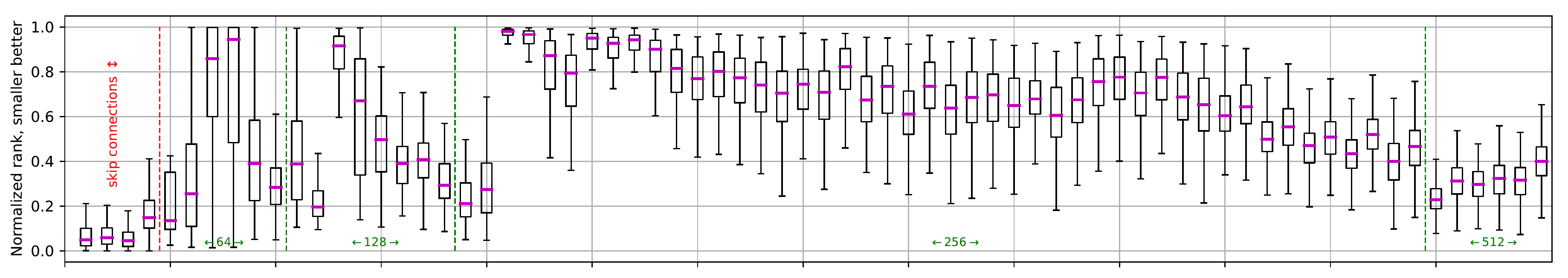}
\includegraphics[width=0.97\textwidth]{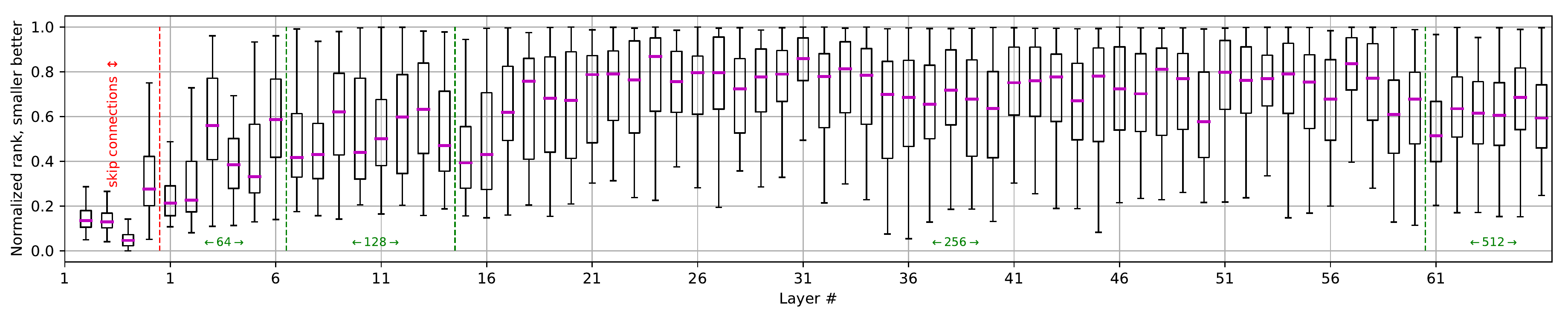}
\caption{Statistics in boxplot form of per-layer ranks before (top) and after (bottom) pruning ResNet-101 with Taylor-FO-BN-50\%. First 4 layers correspond to skip connections, the rest are residual blocks represented by the first 2 convolutional layers per block. We can notice that after pruning most of neurons become more equal than before pruning.}
\label{fig:imagenet_stats}
\end{figure*}

\begin{table}
\centering
\resizebox{0.95\linewidth}{!}
{
\begin{tabular}{lccr}
\toprule
Pruning Method & GFLOPs & Params($10^7$) & $\downarrow$ Error, $\%$ \\ 
\midrule
\midrule
\multicolumn{4}{c}{\textbf{ResNet-101}}  \\
{Taylor-FO-BN-40\% {\bf (Ours)}} & $1.76$ & $1.36$ & $25.84$\\
\cmidrule(l{2em}r{2em}){1-4}
Taylor-FO-BN-50\% {\bf (Ours)} &$\mathbf{2.47}$ & $1.78$ & $\mathbf{24.62}$\\
BN-ISTA v2 \cite{ICLR2018}  &  $3.69$ & $\mathbf{1.73}$ & $25.44$ \\
\cmidrule(l{2em}r{2em}){1-4}
{Taylor-FO-BN-55\% {\bf (Ours)}} & $\mathbf{2.85}$ & $\mathbf{2.07}$ & $\mathbf{24.05}$\\
BN-ISTA v1 \cite{ICLR2018} & $4.47$ & $2.36$ & $24.73$ \\
\cmidrule(l{2em}r{2em}){1-4}
No pruning & $7.80$ & $4.47$ & $\mathbf{22.63}$ \\
{Taylor-FO-BN-75\% {\bf (Ours)}} &$\mathbf{4.70}$ & $\mathbf{3.12}$ & $22.65$ \\
\midrule
\multicolumn{4}{c}{pruning only skip connections}  \\
\midrule
{Taylor-FO-BN-52\% {\bf (Ours)}} & $6.57$ & $3.60$ & $22.94$ \\
{Taylor-FO-BN-22\% {\bf (Ours)}} & $5.19$ & $2.86$& $24.77$ \\
\midrule
\multicolumn{4}{c}{\textbf{ResNet-50}} \\
Taylor-FO-BN-56\% {\bf (Ours)} & $1.34$ & $\mathbf{0.79}$ & $\mathbf{28.31}$ \\
Taylor-FO-BN-56\% (No skip) & $1.28$ & $0.85$ & $30.74$ \\
ThiNet-30~\cite{luo2017thinet} & $\approx\!\mathbf{1.17}$ & $0.87$ & $31.58$ \\
\cmidrule(l{2em}r{2em}){1-4}
Taylor-FO-BN-72\% {\bf (Ours)} & $\mathbf{2.25}$ & $\mathbf{1.42}$ & $\mathbf{25.50}$ \\
NISP-50-B~\cite{yu2017nisp} & $\approx\!2.29$ & $1.43$ & $27.93$ \\ 
ThiNet-70~\cite{luo2017thinet} & $\approx\!2.58$ & $1.69$ & $27.96$ \\
\cmidrule(l{2em}r{2em}){1-4}
Taylor-FO-BN-81\% {\bf (Ours)} & $\mathbf{2.66}$ & $\mathbf{1.79}$ & $\mathbf{24.52}$ \\
SSS~\cite{huang2017data}, ResNet-32 & $2.82$ & $1.86$ & $25.82$ \\
NISP-50-A~\cite{yu2017nisp} & $\approx\!2.97$ & $1.86$ & $27.25$\\ 
\cmidrule(l{2em}r{2em}){1-4}
Taylor-FO-BN-91\% {\bf (Ours)}  & $\mathbf{3.27}$ & $\mathbf{2.26}$ & $\mathbf{23.57}$ \\
No pruning & $4.09$ & $2.56$ & $23.82$ \\
SSS~\cite{huang2017data}, ResNet-41 & $3.47$ & $2.53$ & $24.56$ \\
\cmidrule(l{2em}r{2em}){1-4}
\multicolumn{4}{c}{\textbf{ResNet-34}} \\
No pruning & $3.64$ & $2.18$ & $26.69$ \\
Taylor-FO-BN-82\% {\bf (Ours)} & $2.83$ & $\mathbf{1.72}$ & $\mathbf{27.17}$ \\
Li \etal.~\cite{li2016pruning} & $\mathbf{2.76}$ & $1.93$ & $27.80$ \\

\midrule
\multicolumn{4}{c}{\textbf{VGG11-BN}}  \\
No pruning & $7.61$ & $13.29$ & $\mathbf{29.16}$ \\
{Taylor-FO-BN-50\% \textbf{(Ours)}} & $\mathbf{6.93}$ & $\mathbf{3.18}$ & $29.35$ \\
From scratch \cite{ICLR2019} & $\approx\!\mathbf{6.93}$ & $\approx\!\mathbf{3.18}$ & $30.00$ \\
Slimming \cite{liu2017learning}, from \cite{ICLR2019} & $\approx\!\mathbf{6.93}$ &$\approx\!\mathbf{3.18}$ & $31.38$ \\
\midrule
\multicolumn{4}{c}{\textbf{DenseNet-201}}  \\
No pruning & $4.29$ & $2.20$ & $\mathbf{23.20}$\\
{Taylor-FO-BN-60\% \textbf{(Ours)}} & $3.02$ & $1.25$ & $23.49$ \\
{Taylor-FO-BN-36\% {\bf (Ours)}} & $\mathbf{2.21}$ & $0.90$ & $24.72$ \\
No pruning & $2.74$ & $\mathbf{0.76}$ & $25.57$ \\
\bottomrule
\end{tabular}
}
\caption{Pruning results on ImageNet (1-crop validation errors). }

\label{tab:imagemet_pruning}
\vspace{-10pt}
\end{table}

We use the following settings: $4$ GPUs and a batch size of $256$ examples; we optimized using SGD with initial learning rate $0.01$ (or $0.001$, see Sec.~6.2) decayed a factor $10$ every $10$ epochs, momentum set to $0.9$; pruning and fine-tuning run for $25$ epochs total; we report the best validation accuracy observed. Every $30$ mini-batches we remove $100$ neurons until we reach the predefined number of neurons to be pruned, after which we reset the momentum buffer and continue fine-tuning. By setting the percentage of neurons to remain after pruning to be \emph{X}, we get different versions of the final model and refer to them as \emph{Taylor-FO-BN-X}\%.

Comparison of pruning networks on ImageNet by the proposed method and other methods is presented in the Table~\ref{tab:imagemet_pruning}, where we report total number of FLOPs, number of parameters, and the top-1 error rate. Comparison is grouped by network architecture type and the number of parameters. For ResNet-101 we observe smaller error rates and fewer GFLOPs (by at least $1.22$ GFLOPs) when compare to BN-ISTA method~\cite{ICLR2018}. Pruning only skip connections shows larger errors however makes the final network faster (see Sec~6.2). By pruning $40\%$ of FLOPs and $30\%$ of parameters from original ResNet-101 we only lose $0.02\%$ in accuracy. Pruning results on ResNet-50 and ResNet-34 demonstrate significant improvements over other methods. Additionally we study our method without pruning skip connections, marked as ``No skip'' and observe accuracy loss. Comparison per layer ranking of different layers in ResNet-101 before and after pruning is shown in Fig.~\ref{fig:imagenet_stats}.

\noindent{\bf Pruning neurons with a single step.} As an alternative to iterative pruning, we performed pruning of $10000$ neurons with a single step after $3000$ mini-batches, followed by fine-tuning. This gave a top-1 error of $25.3\%$ 
, which is $0.68$\% higher than \emph{Taylor-FO-BN-50\%}, again emphasizing the benefit of re-evaluating the criterion between pruning iterations.

\noindent{\bf Pruning other networks.}
We also prune the VGG11-BN and DenseNet networks. The former is a simple feed-forward architecture, without skip connections. We prune $50\%$ of neurons across all layers, as per prior work~\cite{ICLR2019,liu2017learning}. Our approach shows only $0.19\%$ loss in accuracy after removing $76\%$ of parameters and improves on the previously reported results by $0.65\%$~\cite{ICLR2019} and more than $2\%$~\cite{liu2017learning}.
DeseNets reuse feature maps multiple times, potentially making them less amenable to pruning. We prune DenseNet-201 and observe that with the same number of FLOPs (\emph{Taylor-FO-BN-52\%}) as DenseNet-121, we have $1.79\%$ lower error.

\section{Conclusions}
\label{sec:conclusions}
In this work, we have proposed a new method for estimating the contribution of a neuron using the Taylor expansion applied on a squared change in loss induced by removing a chosen neuron.
We demonstrated that even the first-order approximation shows significant agreement with true importance, and outperforms prior work on a range of deep networks.
After extensive analysis, we showed that applying the first-order criterion after batch-norms yields the best results, under practical computational and memory constraints.

{\small
\bibliographystyle{ieee}
\bibliography{pruning_biblio}

\begin{thebibliography}{10}\itemsep=-1pt

\bibitem{chauvin1989back}
Y.~Chauvin.
\newblock A back-propagation algorithm with optimal use of hidden units.
\newblock In {\em NIPS}, 1989.

\bibitem{cover2012elements}
T.~M. Cover and J.~A. Thomas.
\newblock {\em Elements of information theory}.
\newblock John Wiley \& Sons, 2012.

\bibitem{daubechies2004iterative}
I.~Daubechies, M.~Defrise, and C.~De~Mol.
\newblock An iterative thresholding algorithm for linear inverse problems with
  a sparsity constraint.
\newblock {\em Communications on Pure and Applied Mathematics}, 2004.

\bibitem{frankle2018lottery}
J.~Frankle and M.~Carbin.
\newblock The lottery ticket hypothesis: Training pruned neural networks.
\newblock {\em arXiv preprint arXiv:1803.03635}, 2018.

\bibitem{gordon2018morphnet}
A.~Gordon, E.~Eban, O.~Nachum, B.~Chen, H.~Wu, T.-J. Yang, and E.~Choi.
\newblock Morphnet: Fast \& simple resource-constrained structure learning of
  deep networks.
\newblock In {\em CVPR}, 2018.

\bibitem{han2016dsd}
S.~Han, J.~Pool, S.~Narang, H.~Mao, S.~Tang, E.~Elsen, B.~Catanzaro, J.~Tran,
  and W.~J. Dally.
\newblock Dsd: regularizing deep neural networks with dense-sparse-dense
  training flow.
\newblock {\em arXiv preprint arXiv:1607.04381}, 2016.

\bibitem{han2015learning}
S.~Han, J.~Pool, J.~Tran, and W.~Dally.
\newblock Learning both weights and connections for efficient neural network.
\newblock In {\em NIPS}, 2015.

\bibitem{hanson1989comparing}
S.~J. Hanson and L.~Y. Pratt.
\newblock Comparing biases for minimal network construction with
  back-propagation.
\newblock In {\em NIPS}, 1989.

\bibitem{he2016deep}
K.~He, X.~Zhang, S.~Ren, and J.~Sun.
\newblock Deep residual learning for image recognition.
\newblock In {\em CVPR}, 2016.

\bibitem{he2016identity}
K.~He, X.~Zhang, S.~Ren, and J.~Sun.
\newblock Identity mappings in deep residual networks.
\newblock In {\em ECCV}, 2016.

\bibitem{he2018progressive}
Y.~He, X.~Dong, G.~Kang, Y.~Fu, and Y.~Yang.
\newblock Progressive deep neural networks acceleration via soft filter
  pruning.
\newblock {\em arXiv preprint arXiv:1808.07471}, 2018.

\bibitem{he2018adc}
Y.~He and S.~Han.
\newblock Adc: Automated deep compression and acceleration with reinforcement
  learning.
\newblock {\em arXiv preprint arXiv:1802.03494}, 2018.

\bibitem{he2017channel}
Y.~He, X.~Zhang, and J.~Sun.
\newblock Channel pruning for accelerating very deep neural networks.
\newblock In {\em ICCV}, 2017.

\bibitem{hinton2015distilling}
G.~Hinton, O.~Vinyals, and J.~Dean.
\newblock Distilling the knowledge in a neural network.
\newblock In {\em arXiv preprint arXiv:1503.02531}, 2015.

\bibitem{huang2017densely}
G.~Huang, Z.~Liu, L.~Van Der~Maaten, and K.~Q. Weinberger.
\newblock Densely connected convolutional networks.
\newblock In {\em CVPR}, 2017.

\bibitem{huang2017data}
Z.~Huang and N.~Wang.
\newblock Data-driven sparse structure selection for deep neural networks.
\newblock {\em arXiv preprint arXiv:1707.01213}, 2017.

\bibitem{ioffe2015batch}
S.~Ioffe and C.~Szegedy.
\newblock Batch normalization: Accelerating deep network training by reducing
  internal covariate shift.
\newblock {\em arXiv preprint arXiv:1502.03167}, 2015.

\bibitem{krizhevsky2009cifar}
A.~Krizhevsky and G.~Hinton.
\newblock Learning multiple layers of features from tiny images.
\newblock {\em Tech Report}, 2009.

\bibitem{krizhevsky2012imagenet}
A.~Krizhevsky, I.~Sutskever, and G.~E. Hinton.
\newblock Imagenet classification with deep convolutional neural networks.
\newblock In {\em NIPS}, pages 1097--1105, 2012.

\bibitem{lebedev2016fast}
V.~Lebedev and V.~Lempitsky.
\newblock Fast convnets using group-wise brain damage.
\newblock In {\em CVPR}, pages 2554--2564, 2016.

\bibitem{lecun1990optimal}
Y.~LeCun, J.~S. Denker, S.~Solla, R.~E. Howard, and L.~D. Jackel.
\newblock Optimal brain damage.
\newblock In {\em NIPS}, 1990.

\bibitem{li2016pruning}
H.~Li, A.~Kadav, I.~Durdanovic, H.~Samet, and H.~P. Graf.
\newblock Pruning filters for efficient convnets.
\newblock {\em ICLR}, 2017.

\bibitem{liu2017learning}
Z.~Liu, J.~Li, Z.~Shen, G.~Huang, S.~Yan, and C.~Zhang.
\newblock Learning efficient convolutional networks through network slimming.
\newblock In {\em ICCV}, 2017.

\bibitem{ICLR2019}
Z.~Liu, M.~Sun, T.~Zhou, G.~Huang, and T.~Darrell.
\newblock Rethinking the value of network pruning.
\newblock {\em arXiv preprint arXiv:1810.05270}, 2018.

\bibitem{louizos2017learning}
C.~Louizos, M.~Welling, and D.~P. Kingma.
\newblock Learning sparse neural networks through $ l\_0 $ regularization.
\newblock {\em arXiv preprint arXiv:1712.01312}, 2017.

\bibitem{luo2017thinet}
J.-H. Luo, J.~Wu, and W.~Lin.
\newblock Thinet: A filter level pruning method for deep neural network
  compression.
\newblock {\em ICCV}, 2017.

\bibitem{ICLR2017}
P.~Molchanov, S.~Tyree, T.~Karras, T.~Aila, and J.~Kautz.
\newblock Pruning convolutional neural networks for resource efficient transfer
  learning.
\newblock {\em ICLR}, 2017.

\bibitem{mozer1989skeletonization}
M.~C. Mozer and P.~Smolensky.
\newblock Skeletonization: A technique for trimming the fat from a network via
  relevance assessment.
\newblock In {\em NIPS}, 1989.

\bibitem{neklyudov2017structured}
K.~Neklyudov, D.~Molchanov, A.~Ashukha, and D.~P. Vetrov.
\newblock Structured bayesian pruning via log-normal multiplicative noise.
\newblock In {\em Advances in Neural Information Processing Systems}, pages
  6775--6784, 2017.

\bibitem{pytorch}
A.~Paszke, S.~Gross, S.~Chintala, G.~Chanan, E.~Yang, Z.~DeVito, Z.~Lin,
  A.~Desmaison, L.~Antiga, and A.~Lerer.
\newblock Automatic differentiation in pytorch.
\newblock In {\em NIPS-W}, 2017.

\bibitem{russakovsky2015imagenet}
O.~Russakovsky, J.~Deng, H.~Su, J.~Krause, S.~Satheesh, S.~Ma, Z.~Huang,
  A.~Karpathy, A.~Khosla, M.~Bernstein, A.~C. Berg, and L.~Fei-Fei.
\newblock {ImageNet Large Scale Visual Recognition Challenge}.
\newblock {\em IJCV}, 2015.

\bibitem{theis2018faster}
L.~Theis, I.~Korshunova, A.~Tejani, and F.~Husz{\'a}r.
\newblock Faster gaze prediction with dense networks and fisher pruning.
\newblock {\em arXiv preprint arXiv:1801.05787}, 2018.

\bibitem{ICLR2018}
J.~Ye, X.~Lu, Z.~Lin, and J.~Z. Wang.
\newblock Rethinking the smaller-norm-less-informative assumption in channel
  pruning of convolution layers.
\newblock {\em ICLR}, 2018.

\bibitem{yu2017nisp}
R.~Yu, A.~Li, C.-F. Chen, J.-H. Lai, V.~I. Morariu, X.~Han, M.~Gao, C.-Y. Lin,
  and L.~S. Davis.
\newblock {NISP}: Pruning networks using neuron importance score propagation.
\newblock {\em CVPR}, 2017.

\bibitem{zhang2016understanding}
C.~Zhang, S.~Bengio, M.~Hardt, B.~Recht, and O.~Vinyals.
\newblock Understanding deep learning requires rethinking generalization.
\newblock {\em ICLR}, 2016.

\end{thebibliography}
}

\clearpage

\section{Supplementary material}
\label{sec:supplementary}
In supplementary material we show additional experimental results on ResNet20 with CIFAR10, ResNet101 on ImageNet. Additionally, we evaluate inference speed of pruned ResNet101 models.

\subsection{ResNet20 on CIFAR10}
We experiment on ResNet20 trained on CIFAR10 in order to compare with the work of~\cite{ICLR2018} (referred to as BN-ISTA) and to evaluate the effect of the ``pruning paradox'' reported in~\cite{ICLR2019}. Our setup is as follows: initial model was trained for $200$ epochs with learning rate $0.1$ and decay by $10$ after $80$ epochs. Final model obtained $92\%$ on the test split and we picked it as an initial model for pruning. 
Pruning and fine-tuning setup is: initial learning rate of $0.1$, decayed by $10$ every $20$ epochs for a total number of $70$ epochs.
While pruning, we remove $10$ neurons every $30$ mini-batches until the predefined number of pruned neurons is reached.

\begin{table*}[ht!]
\centering
\resizebox{\linewidth}{!} 
{
\begin{tabular}{lcccccc|cc}
\toprule
 Strategy & Neurons & BN-ISTA \cite{ICLR2018} & Random & Oracle & Weight magnitude & OBD~\cite{lecun1990optimal} & Taylor FO (Ours)  & Taylor SO (Ours)  \\
\midrule
\midrule
Prune A & 223($\approx 70\%$)  & 90.9\% & 88.22($\pm$0.51) &91.61($\pm$0.10) &86.93($\pm$0.25) &91.57 ($\pm$0.15) & 91.52 ($\pm$0.11) & 91.56 ($\pm$0.14)  \\
Prune A - train from scratch & 223($\approx 70\%$) & & 86.28($\pm$3.59) &89.55($\pm$0.22) &80.97($\pm$4.07) &89.62($\pm$0.24) &89.56($\pm$0.19) &89.63($\pm$0.20) \\
\midrule
Prune B & 119($\approx 35\%$) & 88.8\% & 71.49($\pm$2.35) &89.72($\pm$0.10) &62.03($\pm$1.36)  &89.78 ($\pm$0.16) &89.78 ($\pm$0.18) & 89.76 ($\pm$0.17) \\
Prune B - train from scratch  & 119($\approx 35\%$) & & 77.90($\pm$7.01) &88.25($\pm$0.28) &62.08($\pm$1.08)  &88.14($\pm$0.22) & 88.17($\pm$0.22) & 88.29($\pm$0.19)  \\
\bottomrule
\end{tabular}
}
\caption{Pruning results on ResNet20 for CIFAR10. Only the first layer in every residual block is pruned. Results are averaged over 10 seeds. }
\label{tab:resnet20}
\end{table*}

Results of pruning and training from scratch are summarized in the Table~\ref{tab:resnet20}. We observe that pruning with Random or magnitude based criteria results in the worst performance, primary because they introduce uncorrelated bias to the estimate, we also observe that these 2 methods can be affected by "pruning paradox" as their difference is within a standard deviation of experiments. Our proposed method that relies on estimating feature importance with Taylor expansion of first and second orders outperform BT-ISTA\cite{ICLR2018}. The difference between Optimal Brain Damage and Our methods is not large and within a single standard deviation. We conclude that first order Taylor expansion applied to the gates after BN is a reasonable choice for residual network. It is not affected by ''prunung paradox'' discovered in \cite{ICLR2019}.

\subsection{Additional details on ResNets pruning}
\label{sec:sup_single_step}
Our method can be applied with various pruning scheduling. The scheduling we apply in the paper, named here as \textit{iterative}, removes 100 neurons per every 30 mini-batch updates until we reach predefined number of neurons. Also, all neurons can be removed at once, named as \textit{pruning with a single step}. One more setting, named as \textit{continuous}, prunes 100 neurons every 30 mini-batches only if the training loss is above the predefined threshold (we set it to~1.04). 

\begin{figure}
\centering
\includegraphics[width=\columnwidth,clip]{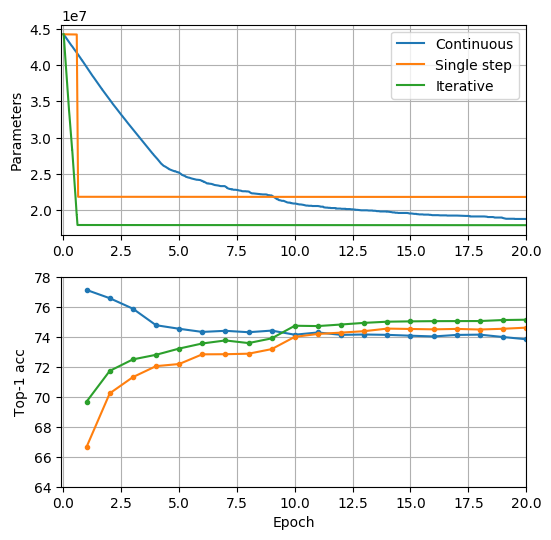}
\caption{Pruning ResNet-101 on Imagenet with 3 different settings. }
\label{fig:imagenet_3mode}
\end{figure}

Progress of ResNet-101 pruning on ImageNet with 3 different pruning scheduling settings is illustrated in Fig. \ref{fig:imagenet_3mode}. All settings had the maximum number of neurons to be pruned as 10000 out of 20096, and the \textit{Iterative} corresponds to \textit{TaylorFO-BN-50}\% in the main paper. Iterative pruning clearly outperforms other settings over all epochs. 

\paragraph{Finetuning details on ImageNet dataset.} When a large number of neurons are removed we found that starting with the larger learning rate works better. Therefore we use starting learning of $0.01$ for the following pruning models: ResNet-101 ( Taylor-FO-BN-40\% and Taylor-FO-BN-22\%), ResNet-50 (Taylor-FO-BN-56\%). All other networks were finetuned with initial learning of $0.001$. Weight decay is set to $0.0$ during finetuning.

\begin{table}[ht!]
\centering
\resizebox{0.99\linewidth}{!}
{
\begin{tabular}{lcccll}
\toprule
Pruning Method & GFLOPs & Params($10^7$) & $\downarrow$ Error, $\%$ & Time, B16 & Time, B256\\ 
\midrule
No pruning & 7.80 & 4.47 & 22.63 & 29.0 & 379.8 \\
{Taylor-FO-BN-75\%} & 4.70 & 3.12 & 22.65 & 24.1 & 313.5\\
{Taylor-FO-BN-55\%} & 2.85 & 2.07 & 24.05 & 21.6 & 261.7\\
{Taylor-FO-BN-50\%} & 2.47 & 1.78 & 24.62 & 20.9 & 251.4\\
{Taylor-FO-BN-40\%} & 1.76 & 1.36 & 25.84 & 21.0 & 223.4\\
\midrule
\multicolumn{6}{c}{pruning only skip connections}  \\
\midrule
{Taylor-FO-BN-52\%} & 6.57 & 3.60 & 22.94 & 25.3 & 326.7\\
{Taylor-FO-BN-22\%} & 5.19 & 2.86& 24.77  & 19.5 & 239.3\\

\bottomrule
\end{tabular}
}
\caption{Batch inference time of models obtained by pruning ResNet-101, time is measured on NVIDIA Tesla V100 in ms with different batch sizes.}

\label{tab:imagemet_speed}
\end{table}

\paragraph{Inference speed.} The main reason behind filter level pruning is computation cost reduction. We evaluate inference time of pruned models in the Table  \ref{tab:imagemet_speed}. Pruning results in inference speed reduction, especially for the larger batch size. Pruning skip connections results in higher time reduction compared to pruning all layers. For example, only by removing 33\% of FLOPs result in 1.59$\times$ speed up of \textit{Taylor-FO-BN-22\%}, while by removing 68\% of FLOPs results only in 1.51$\times$ speed up of \textit{Taylor-FO-BN-50\%}. 

\subsection{Oracle computation details}

Oracle for Table~\ref{tab:imagemet_correlation} is computed from the training set (as~\cite{ICLR2017}) with Eq. ~(\ref{eq:importance}). To check if correlation study is representative we compute the Oracle from the test set. Correlation between Oracles computed on training and testing sets, respectively, is $95.67\%$. After recomputing Table~\ref{tab:imagemet_correlation} using the test set, we observed little change (avg. deviation of $0.04$ between raw table entries) and no reordering of the methods.

\end{document}